\documentclass{article}

\usepackage{amssymb}
\usepackage{centernot}
\usepackage{pifont}
\usepackage{tcolorbox}
\usepackage{rotating}
\tcbuselibrary{breakable}
\usepackage[preprint]{neurips_2026}

\usepackage[utf8]{inputenc} 
\usepackage[T1]{fontenc}    
\usepackage{hyperref}       
\usepackage{url}            
\usepackage{booktabs}       
\usepackage{amsfonts}       
\usepackage{amssymb}        
\usepackage{amsmath}        
\usepackage{nicefrac}       
\usepackage{microtype}      
\usepackage{xcolor}         
\usepackage{graphicx}
\usepackage{subcaption}
\usepackage{svg}
\usepackage{algorithm,algorithmic}
\usepackage{amsthm}

\usepackage{multirow}
\usepackage{tikz}
\usepackage{xcolor}
\usepackage{calc}
\usepackage{fancyvrb}
\usepackage{pifont}

\definecolor{wandbgold}{HTML}{FFBE00}

\theoremstyle{definition}

\theoremstyle{remark}

\definecolor{creamfill}{RGB}{255,249,235}   
\definecolor{creamstroke}{RGB}{210,180,140} 

\title{Continuum Attention}

\author{%
  Afshin Khadangi\\
  SnT, University of Luxembourg\\
  \texttt{afshin.khadanki@uni.lu}\\
  \href{https://afshin.xyz}{\texttt{afshin.xyz}}
}

\begin{document}

\maketitle


\begin{minipage}{\linewidth}
    \centering
    \includegraphics[
        width=\linewidth,
        keepaspectratio
    ]{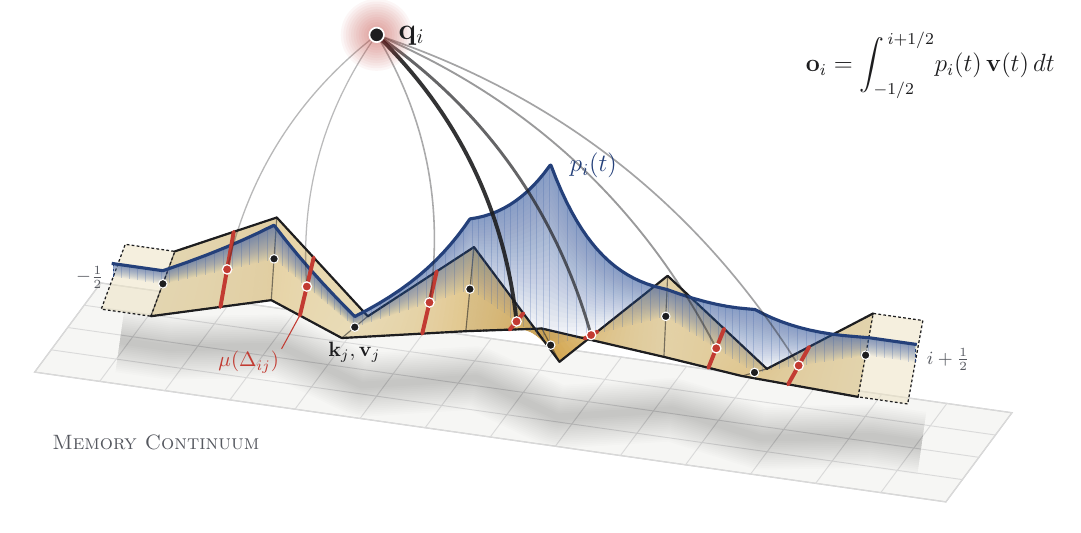}

    \captionof{figure}{
        Overview of Continuum Attention.
        Adjacent key and value representations define a
        continuous memory trajectory. Each query induces
        a relevance density $p_i(t)$ and determines
        adaptive retrieval positions $\mu(\Delta_{ij})$
        within memory segments. Exact integration,
        including the half token boundary extensions,
        produces the attention output.
    }
    \label{fig:ca_overview}
\end{minipage}

\vspace{0.8em}

\begin{abstract}
Conventional causal attention aggregates values at discrete memory positions. We introduce \textbf{\textsc{Continuum Attention (CA)}}, in which neighboring key and value vectors define a continuous, piecewise linear memory trajectory and each query retrieves information by normalized integration over its visible prefix. The segment integrals have closed forms in exponential moments, yielding analytically differentiable attention without numerical quadrature. The resulting operator preserves causal masking, normalized nonnegative effective weights, and the standard Transformer decoder structure. Controlled byte level experiments favor exact continuous integration over matched softmax attention. We additionally evaluate data efficiency, out of domain language modeling, zero-shot multiple choice prediction, robustness, and multilingual generalization. In multilingual evaluation, \textbf{\textsc{CA} consistently outperforms all baselines across every evaluated language}, achieving an \textbf{average 5.8\% reduction in bits per byte (BPB)} relative to the strongest baseline.
\end{abstract}

\section{Introduction}
\label{sec:introduction}
The Transformer has made content-dependent retrieval a central operation in modern sequence modeling \cite{bahdanau2015neural,vaswani2017attention}. Standard attention interprets the stored keys and values as a discrete set of memory locations. For a query $\mathbf q_i$, a causal attention head forms scores $s_{ij}=\gamma\mathbf q_i^\top\mathbf k_j/\sqrt{d_h}$ and aggregates $\mathbf o_i=\sum_{j\le i}\operatorname{softmax}(s_{i,:})_j\mathbf v_j$. The queries and representations are continuous vectors, but the retrieval operator only evaluates memory at the token indices. This distinction raises a natural question: can the representation learned \emph{between} adjacent memory locations contribute directly to retrieval?

We investigate a precise answer through \emph{Continuum Attention} (CA). Neighboring keys and values form piecewise linear trajectories $\mathbf k(t)$ and $\mathbf v(t)$. At query position $i$, a normalized exponential relevance density integrates the continuous value curve over the visible domain $[-\tfrac12,i+\tfrac12]$; constant half-token endpoint extensions provide a consistent discrete-mass convention. Since the score is affine inside each interval, forward integration is exact via the zeroth and first exponential moments, and analytic score derivatives require only the second moment. This is a continuous retrieval operator with a simple equivalent representation as normalized nonnegative weights on the original value vectors.

Other departures from the discrete softmax rule include sparse probability mappings, adaptive attention spans, and fast kernelized attention \cite{martins2016sparsemax,sukhbaatar2019adaptive,choromanski2021performer}. These ideas address distinct tradeoffs in sparsity, receptive field, and computational cost.

Continuous attention has been studied through probability distributions and operator theoretic formulations. CA advances this direction with exact causal integration over piecewise linear key and value trajectories. Sparse and continuous attention distributions were developed by Martins and colleagues \cite{martins2020sparsecontinuous,martins2022sparsecontinuous}, and a measure theoretic treatment of attention as an integral operator was given by Vuckovic et al.\ \cite{vuckovic2021regularity}. Particularly relevant is the prior work titled \emph{Continuum Attention for Neural Operators} by Calvello et al.\ \cite{calvello2025continuum}, which places attention in function spaces for operator learning. Our contribution is an exactly integrable causal attention operator over token-indexed, piecewise linear key and value trajectories, with analytical gradients and autoregressive language modeling evaluations. Unlike quadrature based discretizations of a general function space attention operator, the segment formulas here require no numerical integration.

We evaluate CA at two scales. First, controlled small-model experiments hold the architecture and training data fixed while varying the attention operator and score gain, using five seeds, two context lengths, and fixed and learned gain controls. Second, we train a 101M-parameter causal decoder on 26 billion FineWeb-Edu bytes using four NVIDIA V100 GPUs. Three-seed evaluations compare CA against approximately parameter matched baselines across web text, books, source code, multilingual text, zero-shot multiple-choice tasks, and byte level perturbations.

Our contributions are (i) an exact normalized causal integral operator over interpolated memory trajectories; (ii) its stable finite moment forward computation and analytic backward equations; (iii) controlled attention temperature comparisons at small scale and a substantially larger byte level model study; and (iv) a GPU implementation and verification protocol covering numerical correctness, causal independence, memory use, and practical throughput.

\section{Method}
\label{sec:method}

We formulate Continuum Attention (CA) as an exact integral
operator over a continuous memory trajectory constructed from
the key and value representations of a causal Transformer.
The construction is applied independently within each attention
head and Transformer layer. For clarity, we omit the layer and
head indices throughout the derivation.

\subsection{Continuous Memory Representation}
\label{sec:continuous_memory}

Consider a sequence of $T$ token representations projected
into queries, keys, and values:
\begin{equation}
    \mathbf{q}_i = W_Q\mathbf{x}_i,
    \qquad
    \mathbf{k}_i = W_K\mathbf{x}_i,
    \qquad
    \mathbf{v}_i = W_V\mathbf{x}_i,
\end{equation}
where $\mathbf{q}_i,\mathbf{k}_i,\mathbf{v}_i
\in\mathbb{R}^{d_h}$ and $d_h$ denotes the dimension
of an attention head.

For a causal query at position $i$, only memory positions
$j\leq i$ are accessible. We construct continuous key and
value functions by interpolating adjacent memory vectors.
For each interval $[j-1,j]$, where $1\leq j\leq i$, let
$u=t-(j-1)\in[0,1]$ and define
\begin{align}
    \mathbf{k}_j(u)
    &=
    (1-u)\mathbf{k}_{j-1}
    +u\mathbf{k}_j,
    \label{eq:ca_key_curve}
    \\
    \mathbf{v}_j(u)
    &=
    (1-u)\mathbf{v}_{j-1}
    +u\mathbf{v}_j.
    \label{eq:ca_value_curve}
\end{align}

These segments form continuous functions
$\mathbf{k}(t)$ and $\mathbf{v}(t)$ over $[0,i]$.
We extend both functions constantly over the intervals
$[-1/2,0]$ and $[i,i+1/2]$, using the first and last
visible memory vectors, respectively.

The resulting memory representation preserves the original
token ordering and causal visibility while providing
well defined representations at every position along
the memory trajectory.

\subsection{Query Adaptive Relevance Density}
\label{sec:query_adaptation}

Each query induces a relevance score over the continuous
memory domain:
\begin{equation}
    s_i(t)
    =
    \frac{\gamma}{\sqrt{d_h}}
    \mathbf{q}_i^\top\mathbf{k}(t),
    \label{eq:ca_continuous_score}
\end{equation}
where $\gamma>0$ is the attention score gain.
For a fixed attention temperature $\tau$,
$\gamma=1/\tau$.

Within segment $j$, define the endpoint scores
\begin{equation}
    a_{ij}=s_i(j-1),
    \qquad
    b_{ij}=s_i(j),
    \qquad
    \Delta_{ij}=b_{ij}-a_{ij}.
\end{equation}

Since the key trajectory is linear within the segment,
the score is affine:
\begin{equation}
    s_{ij}(u)
    =
    a_{ij}+u\Delta_{ij},
    \qquad
    u\in[0,1].
    \label{eq:ca_affine_score}
\end{equation}

The query therefore induces an exponential relevance density
\begin{equation}
    p_i(t)
    =
    \frac{\exp(s_i(t))}{Z_i},
    \qquad
    Z_i
    =
    \int_{-1/2}^{i+1/2}
    \exp(s_i(t))\,dt.
    \label{eq:ca_density}
\end{equation}

CA retrieves information by integrating the continuous
value trajectory against this density:
\begin{equation}
    \boxed{
    \mathbf{o}_i
    =
    \int_{-1/2}^{i+1/2}
    p_i(t)\mathbf{v}(t)\,dt
    }.
    \label{eq:ca_operator}
\end{equation}

The adaptation depends directly on the query and the geometry
of neighboring keys. In particular,
\begin{equation}
    \Delta_{ij}
    =
    \frac{\gamma}{\sqrt{d_h}}
    \mathbf{q}_i^\top
    (\mathbf{k}_j-\mathbf{k}_{j-1}),
    \label{eq:ca_query_delta}
\end{equation}
so different queries induce different relevance profiles
over the same memory segments. Distinct learned projections
across layers and heads further provide different memory
trajectories and retrieval behavior.

\subsection{Exact Segment Integration}
\label{sec:exact_integration}

The affine score in Eq.~\eqref{eq:ca_affine_score}
allows the continuous integral to be evaluated exactly.
Define the exponential moments
\begin{equation}
    J_n(\Delta)
    =
    \int_0^1
    u^n\exp(\Delta u)\,du,
    \qquad n\in\{0,1,2\}.
    \label{eq:ca_moments}
\end{equation}

For $\Delta\neq 0$, these moments admit the expressions
\begin{align}
    J_0(\Delta)
    &=
    \frac{e^\Delta-1}{\Delta},
    \label{eq:ca_j0}
    \\
    J_1(\Delta)
    &=
    \frac{(\Delta-1)e^\Delta+1}{\Delta^2},
    \label{eq:ca_j1}
    \\
    J_2(\Delta)
    &=
    \frac{
        (\Delta^2-2\Delta+2)e^\Delta-2
    }{\Delta^3}.
    \label{eq:ca_j2}
\end{align}
Their continuous limits at $\Delta=0$ are
$J_0(0)=1$, $J_1(0)=1/2$, and $J_2(0)=1/3$.

For a segment connecting positions $j-1$ and $j$,
its unnormalized probability mass is
\begin{align}
    M_{ij}
    &=
    \int_0^1
    \exp(a_{ij}+u\Delta_{ij})\,du
    \nonumber\\
    &=
    e^{a_{ij}}J_0(\Delta_{ij}).
    \label{eq:ca_segment_mass}
\end{align}

Its contribution to the value numerator is
\begin{align}
    \mathbf{U}_{ij}
    &=
    \int_0^1
    \exp(a_{ij}+u\Delta_{ij})
    \mathbf{v}_j(u)\,du
    \nonumber\\
    &=
    e^{a_{ij}}
    \Big[
        \big(J_0(\Delta_{ij})
             -J_1(\Delta_{ij})\big)
        \mathbf{v}_{j-1}
    \nonumber\\
    &\hspace{37mm}
        +J_1(\Delta_{ij})\mathbf{v}_j
    \Big].
    \label{eq:ca_segment_numerator}
\end{align}

Thus, the contribution of each memory segment is obtained
from the endpoint scores and values using a finite number
of analytical operations.

\paragraph{Adaptive retrieval positions.}

The normalized mean interpolation coordinate within segment
$j$ is
\begin{equation}
    \mu(\Delta_{ij})
    =
    \frac{J_1(\Delta_{ij})}{J_0(\Delta_{ij})}.
    \label{eq:ca_mu}
\end{equation}

For $\Delta\neq 0$,
\begin{equation}
    \mu(\Delta)
    =
    \frac{1}{1-e^{-\Delta}}
    -\frac{1}{\Delta},
    \qquad
    \mu(0)=\frac12.
    \label{eq:ca_mu_closed}
\end{equation}

Consequently,
\begin{equation}
    \mathbf{U}_{ij}
    =
    M_{ij}
    \Big[
        (1-\mu(\Delta_{ij}))\mathbf{v}_{j-1}
        +\mu(\Delta_{ij})\mathbf{v}_j
    \Big].
    \label{eq:ca_adaptive_value}
\end{equation}

The function $\mu(\Delta)$ increases monotonically from
zero to one. Positive score differences shift the effective
retrieval position toward the right endpoint, while negative
differences shift it toward the left endpoint. Equal scores
produce the midpoint of the segment.

This gives CA two complementary forms of query adaptation.
The segment mass $M_{ij}$ determines how strongly a memory
interval contributes, while $\mu(\Delta_{ij})$ determines
where information is retrieved within that interval.
Both quantities are determined analytically for each query.

\subsection{Boundary Contributions and Normalization}
\label{sec:ca_boundaries}

To treat endpoint memory positions consistently, we include
constant half unit extensions at both ends of the visible
memory domain. Their contributions to the normalizer and
value numerator are
\begin{align}
    Z_i^{\mathrm{cap}}
    &=
    \frac12 e^{s_i(0)}
    +\frac12 e^{s_i(i)},
    \\
    \mathbf{U}_i^{\mathrm{cap}}
    &=
    \frac12 e^{s_i(0)}\mathbf{v}_0
    +\frac12 e^{s_i(i)}\mathbf{v}_i.
\end{align}

The complete normalization factor and numerator are
\begin{align}
    Z_i
    &=
    Z_i^{\mathrm{cap}}
    +\sum_{j=1}^{i}M_{ij},
    \label{eq:ca_total_mass}
    \\
    \mathbf{U}_i
    &=
    \mathbf{U}_i^{\mathrm{cap}}
    +\sum_{j=1}^{i}\mathbf{U}_{ij}.
    \label{eq:ca_total_numerator}
\end{align}

The final attention output is therefore
\begin{equation}
    \boxed{
    \mathbf{o}_i
    =
    \frac{
        \mathbf{U}_i^{\mathrm{cap}}
        +\displaystyle\sum_{j=1}^{i}\mathbf{U}_{ij}
    }{
        Z_i^{\mathrm{cap}}
        +\displaystyle\sum_{j=1}^{i}M_{ij}
    }.
    }
    \label{eq:ca_final}
\end{equation}

For $i=0$, the two half unit contributions together yield
the output $\mathbf{o}_0=\mathbf{v}_0$.

The endpoint convention also ensures that when all visible
key scores are equal, each discrete value receives identical
total mass, and Eq.~\eqref{eq:ca_final} reduces exactly to
uniform causal attention.

Because the segment contributions are nonnegative, CA can
equivalently be written as
\begin{equation}
    \mathbf{o}_i
    =
    \sum_{j=0}^{i}
    w_{ij}\mathbf{v}_j,
    \qquad
    w_{ij}\geq 0,
    \qquad
    \sum_{j=0}^{i}w_{ij}=1.
    \label{eq:ca_discrete_equivalence}
\end{equation}

The effective weights combine contributions from adjacent
memory segments and the boundary terms. This representation
allows CA to be integrated into a conventional Transformer
attention block.

\subsection{Analytical Differentiation}
\label{sec:ca_backward}

The forward operator admits exact analytical derivatives.
The second order moment $J_2$ is sufficient because
both the key scores and values are linear within each
memory segment.

Let $\mathcal{L}$ denote the training objective, and
let
\begin{equation}
    \mathbf{g}_i
    =
    \frac{\partial\mathcal{L}}
         {\partial\mathbf{o}_i},
    \qquad
    r_{ij}
    =
    \mathbf{g}_i^\top
    (\mathbf{v}_j-\mathbf{o}_i).
    \label{eq:ca_backward_terms}
\end{equation}

For segment $j$, the contribution to the gradient with
respect to its left endpoint score is
\begin{align}
    \left.
    \frac{\partial\mathcal{L}}
         {\partial s_i(j-1)}
    \right|_{\mathrm{seg}\,j}
    &=
    \frac{e^{a_{ij}}}{Z_i}
    \Big[
       \big(J_0-2J_1+J_2\big)r_{i,j-1}
    \nonumber\\
    &\hspace{18mm}
       +\big(J_1-J_2\big)r_{ij}
    \Big],
    \label{eq:ca_grad_left}
\end{align}
where the moments are evaluated at $\Delta_{ij}$.

The corresponding contribution to the right endpoint is
\begin{align}
    \left.
    \frac{\partial\mathcal{L}}
         {\partial s_i(j)}
    \right|_{\mathrm{seg}\,j}
    &=
    \frac{e^{a_{ij}}}{Z_i}
    \Big[
        \big(J_1-J_2\big)r_{i,j-1}
    \nonumber\\
    &\hspace{18mm}
        +J_2r_{ij}
    \Big].
    \label{eq:ca_grad_right}
\end{align}

Each endpoint receives the sum of the contributions from
its incident segments. The first and last visible scores
also receive the boundary contributions
\begin{align}
    \left.
    \frac{\partial\mathcal{L}}
         {\partial s_i(0)}
    \right|_{\mathrm{cap}}
    &=
    \frac{e^{s_i(0)}}{2Z_i}r_{i0},
    \\
    \left.
    \frac{\partial\mathcal{L}}
         {\partial s_i(i)}
    \right|_{\mathrm{cap}}
    &=
    \frac{e^{s_i(i)}}{2Z_i}r_{ii}.
\end{align}

The value gradients follow from the effective weights in
Eq.~\eqref{eq:ca_discrete_equivalence}:
\begin{equation}
    \frac{\partial\mathcal{L}}
         {\partial\mathbf{v}_j}
    =
    \sum_{i=j}^{T-1}
    w_{ij}\mathbf{g}_i.
    \label{eq:ca_grad_values}
\end{equation}

Writing $g_{ij}^{s}$ for the complete derivative with
respect to $s_i(j)$, the query and key gradients are
\begin{align}
    \frac{\partial\mathcal{L}}
         {\partial\mathbf{q}_i}
    &=
    \frac{\gamma}{\sqrt{d_h}}
    \sum_{j=0}^{i}
    g_{ij}^{s}\mathbf{k}_j,
    \label{eq:ca_grad_query}
    \\
    \frac{\partial\mathcal{L}}
         {\partial\mathbf{k}_j}
    &=
    \frac{\gamma}{\sqrt{d_h}}
    \sum_{i=j}^{T-1}
    g_{ij}^{s}\mathbf{q}_i.
    \label{eq:ca_grad_key}
\end{align}

For the learned gain variant, we parameterize
$\gamma=\exp(\rho)$ independently for each layer and head.
Its gradient is
\begin{equation}
    \frac{\partial\mathcal{L}}{\partial\rho}
    =
    \sum_{i=0}^{T-1}
    \sum_{j=0}^{i}
    g_{ij}^{s}s_i(j).
    \label{eq:ca_grad_gain}
\end{equation}

These expressions provide analytical differentiation
through the complete attention operator using exponential
moments of order at most two.

\subsection{Numerical Evaluation and Complexity}
\label{sec:ca_implementation}

Direct evaluation of exponential moments can lose numerical
precision when endpoint scores are nearly equal. We therefore
evaluate the moment expressions using series expansions near
$\Delta=0$, with
\begin{equation}
    J_n(\Delta)
    =
    \sum_{r=0}^{\infty}
    \frac{\Delta^r}
         {r!(n+r+1)}.
    \label{eq:ca_moment_series}
\end{equation}

For general score differences, we use numerically stable
exponential expressions and scale the integrals relative
to the larger endpoint score. The normalizer is computed
after subtracting the maximum visible score for each
query, preserving the normalized output. 
Efficient and numerically stable evaluation of exponential
normalizers has also been investigated for conventional
softmax attention \cite{milakov2018online}.

All segment integrals can be evaluated independently.
Their contributions are accumulated into the effective
attention weights, followed by multiplication with the
value matrix. This permits parallel GPU implementation
without numerical quadrature or sequential integration
over memory positions.

For a sequence of length $T$ and head dimension $d_h$,
the attention computation requires
$\mathcal{O}(T^2d_h)$ operations per head, matching the
asymptotic complexity of dense softmax attention.
An implementation that materializes the attention matrix
requires $\mathcal{O}(T^2)$ attention storage, while
tiled implementations can accumulate segment contributions
without retaining the entire matrix. The experiments use custom Metal and CUDA kernels for exact
segment integration and analytical score gradients, each checked against
an independent high-precision mathematical reference. The tiled CUDA
implementation materializes intermediate score and weight tiles;
therefore, an efficient implementation need not have linear attention
storage, even though the mathematical operator has no requirement to
materialize a full dense attention matrix.

\subsection{Integration into Causal Transformers}
\label{sec:ca_transformer}

CA replaces the attention operator within each Transformer
block while retaining the surrounding decoder architecture.
Queries, keys, and values are obtained from the usual learned
linear projections, and the head outputs are concatenated
and projected back to the model dimension. Residual
connections, normalization, feedforward layers, positional
embeddings, and the language modeling objective are retained.

We investigate fixed and learned gain configurations in controlled
small model experiments.
The fixed gain variant, denoted CA-P, uses
$\gamma=1/0.30$ in every attention head.
The learned gain variant, denoted CA-H, initializes
each head at the same gain and optimizes $\rho$
during training. Query adaptation through
Eq.~\eqref{eq:ca_query_delta} occurs in both
configurations.

At larger model scale, each head instead uses a learned positive gain
initialized at $\gamma=1$ (effective $\tau=1$), permitting direct
training of the same operator without altering the surrounding decoder.
The fixed gain formulation introduces no additional
trainable parameters relative to a conventional softmax
attention layer with the same projection dimensions.
This enables a direct architectural comparison under
matched model size, attention temperature, and
training budget.

\section{Experiments and Results}
\label{sec:experiments}
We study the operator first in a controlled five-seed small model ablation (see Appendix \ref{app:experimental_setup}), then in a three-seed larger byte level decoder trained on four GPUs.

\subsection{Large Model: Architecture and Training}
\label{sec:large_setup}

\paragraph{Architecture.} We train a byte level decoder with a 256-symbol vocabulary, context length 2,048, width 1,024, eight causal Transformer blocks, 16 attention heads of dimension 64, and feedforward expansion factor four. The network uses RMS normalization, rotary positional representations \cite{su2024roformer}, GELU feedforward activations, residual connections, independent input/output byte embeddings, and no dropout. All 128 head specific log score gains are learned and initialized at zero, corresponding to $\gamma=1$ (effective $\tau=1$). The model has \textbf{101,205,120 trainable parameters}. 

\paragraph{Optimization and compute.} Training streams UTF-8 bytes from FineWeb-Edu \cite{penedo2024fineweb}, predicts the next byte, and uses 50,000 optimizer steps. Four NVIDIA Tesla V100 GPUs run data parallel training; each GPU processes microbatches of eight contexts with eight gradient accumulation steps. The effective batch is $4\times8\times8\times2048=524{,}288$ input bytes per optimizer step, and the recorded total is \textbf{26,214,400,000 training bytes}. We use AdamW \cite{loshchilov2019adamw} with maximum learning rate $3\times10^{-4}$, a 2,000 step warmup, cosine decay to $3\times10^{-5}$, $\beta=(0.9,0.95)$, weight decay 0.1, and gradient norm clipping at 1.0. Training uses FP16 arithmetic and gradient scaling.

\paragraph{Baselines.} We compare against four approximately 100M-parameter byte-level baselines:

\begin{itemize}
\item \textbf{Transformer:} a flat causal Transformer operating on all 2{,}048 byte positions at every layer \citep{vaswani2017attention};
\item \textbf{FastWeight:} a flat model replacing self-attention with gated fast-weight memory blocks \citep{schlag2021linear,yang2024gla};
\item \textbf{BLT:} the Byte Latent Transformer, which compresses bytes into entropy-dependent latent patches \citep{pagnoni2025blt}; and
\item \textbf{AU-Net:} an autoregressive U-Net with hierarchical downsampling, upsampling, and skip connections \citep{videau2025aunet}.
\end{itemize}

All baselines use the same training corpus, byte context length, optimizer-step budget, effective batch size, learning-rate schedule, numerical precision, and hardware. Parameter counts are matched as closely as permitted by each architecture.

\paragraph{Evaluation.} The checkpoints are evaluated in full sequence, teacher forced next byte prediction. BPB is the summed negative log likelihood divided by the number of scored UTF-8 bytes and $\ln2$. Evaluation uses 2,048 byte contexts with a one byte prediction shift and no repetition of exhausted examples. Multiple choice tasks score the length normalized likelihood of each UTF-8 answer candidate; the evaluator left truncates context as needed to preserve candidate suffixes. HellaSwag corruptions modify prompt bytes using deterministic transformations.

\subsection{Training-Loss Dynamics and Data Efficiency}
\label{sec:large_curve}
\begin{figure*}[t]
\centering
\includegraphics[width=\textwidth]{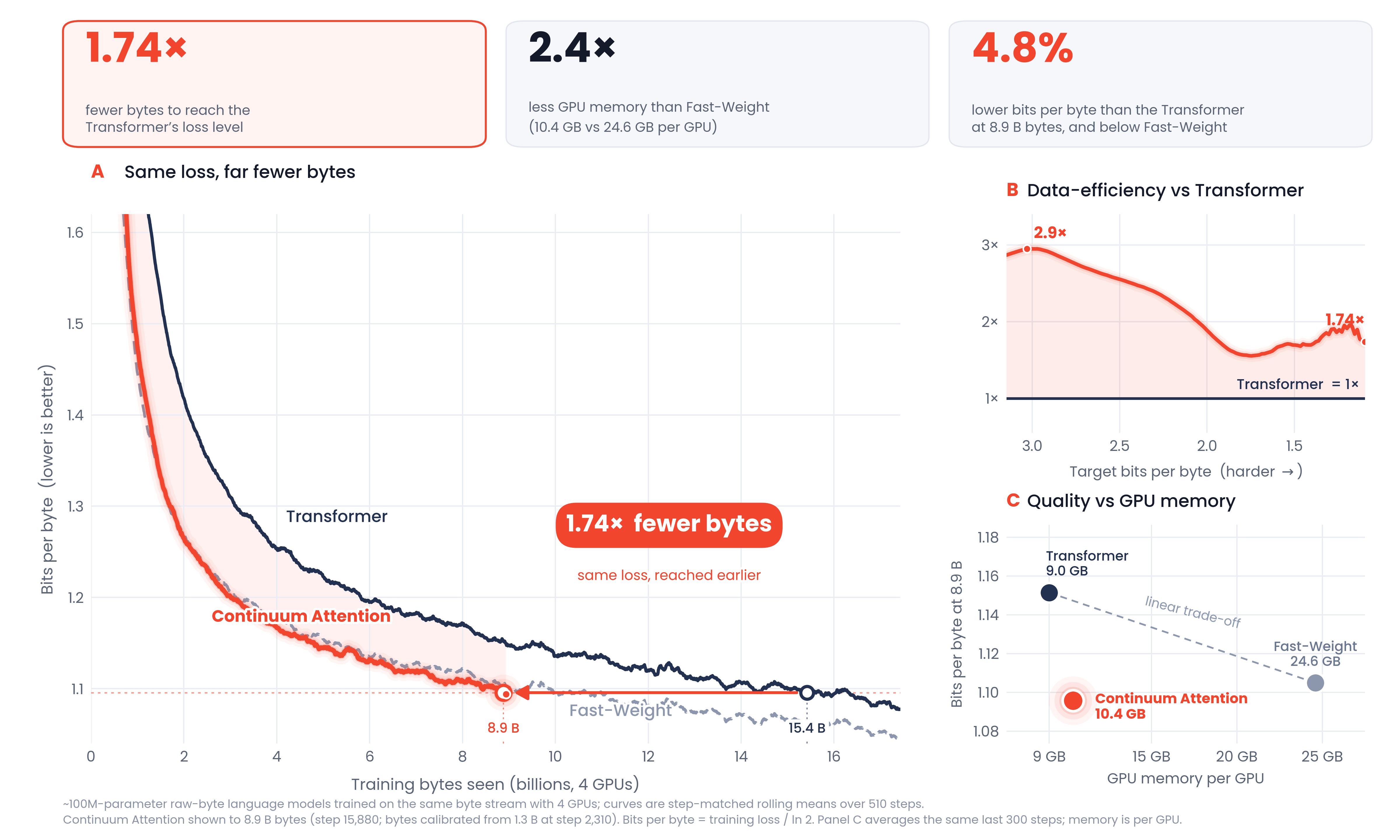}
\caption{100M parameter raw byte training comparison on a shared byte stream with four GPUs. Left: step matched rolling mean of training BPB versus bytes processed for Continuum Attention, a Transformer, and FastWeight. Upper right: interpolated ratio of bytes needed to reach matched training BPB relative to the Transformer. Lower right: observed loss and per GPU allocated memory at a selected training budget.}
\label{fig:training_100m}
\end{figure*}

Figure~\ref{fig:training_100m} compares learning curves on a shared raw-byte stream. At the target of approximately 1.10 training BPB, CA reaches the target after about 8.9 billion training bytes, while the Transformer reaches it after about 15.4 billion bytes. This corresponds to a measured data efficiency ratio of $15.4/8.9\approx1.74$, or about 42\% fewer bytes for the selected threshold. At the 8.9 billion byte comparison point, we report 4.8\% lower BPB than the Transformer.

It also represents approximately 10.4\,GB per GPU for CA, 9.0\,GB for the Transformer, and 24.6\,GB for FastWeight at the selected comparison point. CA uses about $24.6/10.4\approx2.37$ times less GPU memory than FastWeight, while consuming slightly more memory than the Transformer.

\subsection{Language Modeling Across Domains}
\label{sec:large_bpb}

\begin{figure}[t]
    \centering
    \includegraphics[width=\textwidth]{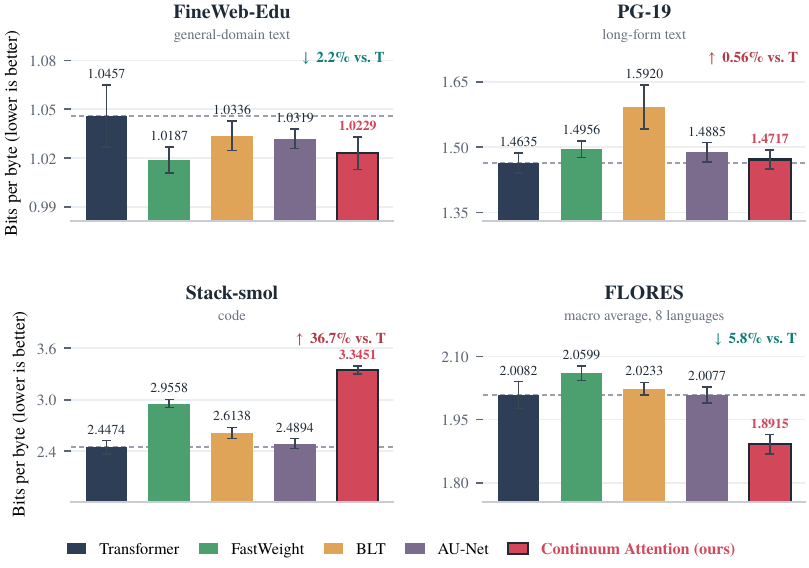}
    \caption{Cross-domain language modeling performance across five architectures, averaged over three seeds. Continuum Attention achieves the lowest FLORES macro-average BPB, while performance across other domains varies by architecture. Lower is better.}
    \label{fig:main_bpb}
\end{figure}

\begin{table}[t]
\centering
\small
\setlength{\tabcolsep}{4.0pt}
\begin{tabular}{lrrrr}
\toprule
Model & FineWeb-Edu & PG-19 & Stack-smol & FLORES \\
\midrule
Transformer & 1.0457 & \textbf{1.4635} & \textbf{2.4474} & 2.0082 \\
FastWeight & \textbf{1.0187} & 1.4956 & 2.9558 & 2.0599 \\
BLT & 1.0336 & 1.5920 & 2.6138 & 2.0233 \\
AU-Net & 1.0319 & 1.4885 & 2.4894 & 2.0077 \\
\midrule
Continuum Attention & 1.0229 & 1.4717 & 3.3451 & \textbf{1.8915} \\
\bottomrule
\end{tabular}
\caption{Language modeling in average bits per byte across three seeds (lower is better). FLORES is the unweighted mean across eight languages.}
\label{tab:large_bpb}
\end{table}

Figure~\ref{fig:main_bpb} and Table~\ref{tab:large_bpb} compare language modeling performance across four evaluation domains. The CA scores 1.0229 BPB on FineWeb-Edu, 1.4717 on PG-19, and 3.3451 on Stack-smol. CA achieves the lowest FLORES macro-average BPB (1.8915 BPB) and competitive performance on FineWeb-Edu and PG-19. Performance varies across domains, with the Transformer achieving the lowest BPB on Stack-smol.

\begin{figure}[t]
    \centering
    \includegraphics[width=\textwidth]{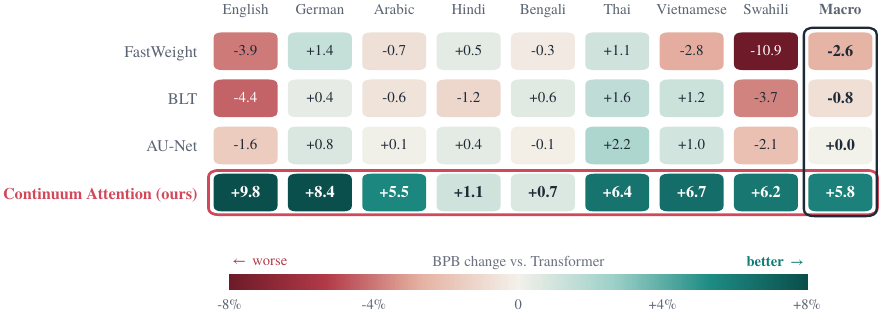}
    \caption{Relative BPB improvement over the Transformer across eight FLORES languages and their macro average. Continuum Attention consistently outperforms all baselines across all eight languages, achieving a 5.8\% average improvement. Positive values indicate better performance.}
    \label{fig:flores_heatmap}
\end{figure}

\subsection{Eight-Language Evaluation}
\label{sec:large_flores}
\begin{table*}[t]
\centering
\scriptsize
\setlength{\tabcolsep}{3.2pt}
\resizebox{\textwidth}{!}{%
\begin{tabular}{lrrrrrrrrr}
\toprule
Model & English & German & Arabic & Hindi & Bengali & Thai & Vietnamese & Swahili & Macro \\
\midrule
Transformer & 1.2013 & 2.4014 & 2.1194 & 1.4265 & 1.6157 & 1.7914 & 2.3884 & 3.1216 & 2.0082 \\
FastWeight & 1.2484 & 2.3674 & 2.1350 & 1.4196 & 1.6205 & 1.7722 & 2.4547 & 3.4613 & 2.0599 \\
BLT & 1.2543 & 2.3916 & 2.1316 & 1.4430 & 1.6065 & 1.7627 & 2.3588 & 3.2376 & 2.0233 \\
AU-Net & 1.2208 & 2.3831 & 2.1172 & 1.4202 & 1.6175 & 1.7526 & 2.3635 & 3.1865 & 2.0077 \\
\midrule
Continuum Attention & \textbf{1.0836} & \textbf{2.1987} & \textbf{2.0025} & \textbf{1.4106} & \textbf{1.6039} & \textbf{1.6771} & \textbf{2.2291} & \textbf{2.9266} & \textbf{1.8915} \\
\bottomrule
\end{tabular}}
\caption{Mean FLORES-200 \texttt{devtest} BPB for eight languages (lower is better) across three seeds; Macro averages languages uniformly.}
\label{tab:large_flores}
\end{table*}

Figure \ref{fig:flores_heatmap} and Table \ref{tab:large_flores} report the multilingual evaluation results. On this suite the CA yields an unweighted macro average of 1.8915 BPB. Individual results range from 1.0836 (English) to 2.9266 (Swahili). \textbf{CA achieves the lowest mean BPB among all five architectures in every evaluated language.}

\subsection{Zero-Shot Multiple Choice and Byte Perturbations}
\label{sec:large_reasoning}

\begin{figure}[t]
    \centering
    \includegraphics[width=\textwidth]{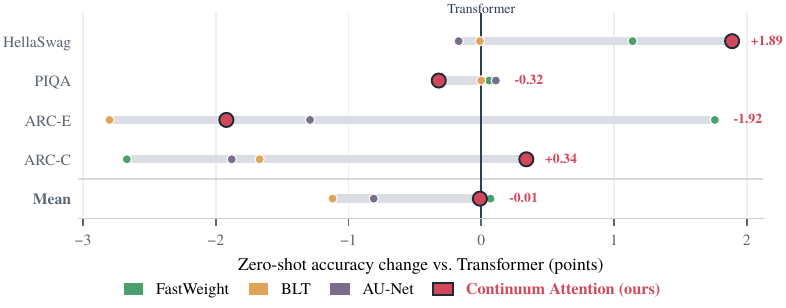}
    \caption{Zero-shot multiple-choice accuracy differences relative to the Transformer across four benchmarks and their mean. Continuum Attention achieves the largest gains on HellaSwag and ARC-Challenge, with competitive overall accuracy. Values represent percentage-point differences.}
    \label{fig:mc}
\end{figure}

\begin{table}[t]
\centering
\small
\setlength{\tabcolsep}{4pt}
\begin{tabular}{lrrrrr}
\toprule
Model & HellaSwag & PIQA & ARC-E & ARC-C & Mean \\
\midrule
Transformer & 28.81 & 60.55 & 35.96 & 22.74 & 37.02 \\
FastWeight & 29.95 & 60.61 & \textbf{37.72} & 20.07 & \textbf{37.09} \\
BLT & 28.80 & 60.55 & 33.16 & 21.07 & 35.90 \\
AU-Net & 28.64 & \textbf{60.66} & 34.67 & 20.86 & 36.21 \\
\midrule
Continuum Attention & \textbf{30.70} & 60.23 & 34.04 & \textbf{23.08} & 37.01 \\
\bottomrule
\end{tabular}
\caption{Mean zero-shot multiple choice accuracy in percent across three seeds (higher is better). Mean is the unweighted four task mean.}
\label{tab:large_mc}
\end{table}

Figure \ref{fig:mc} and Table \ref{tab:large_mc} represent multi-choice evaluation results. The CA reaches 30.70\% HellaSwag, 60.23\% PIQA, 34.04\% ARC-Easy, and 23.08\% ARC-Challenge, averaging approximately 37.01\%. HellaSwag and ARC-Challenge are numerically the highest entries among the model scores, while PIQA and ARC-Easy are not.

\begin{figure}[t]
    \centering
    \includegraphics[width=\textwidth]{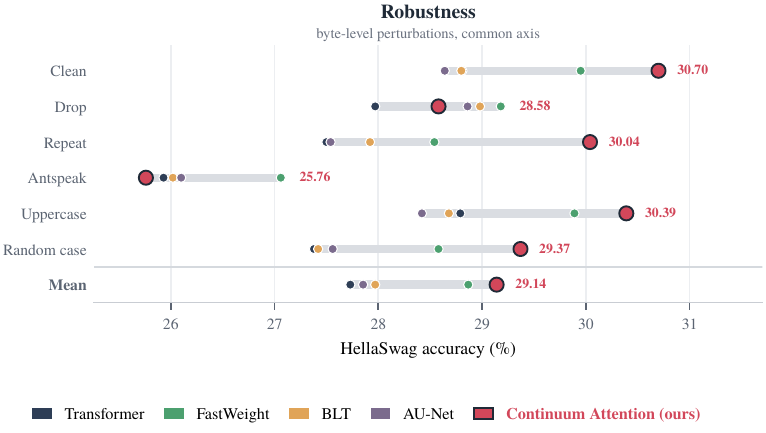}
    \caption{HellaSwag accuracy under deterministic byte level prompt perturbations, averaged over three seeds. Continuum Attention performs strongly under clean, repeated byte, uppercase, and random case conditions, while showing reduced robustness to byte deletion and antspeak transformations. Higher is better.}
    \label{fig:robustness}
\end{figure}

\begin{table*}[t]
\centering
\small
\setlength{\tabcolsep}{6pt}
\begin{tabular}{lrrrrrr}
\toprule
Model & Clean & Drop & Repeat & Antspeak & Uppercase & Randomcase \\
\midrule
Transformer & 28.81 & 27.97 & 27.50 & 25.93 & 28.79 & 27.38 \\
FastWeight & 29.95 & \textbf{29.18} & 28.54 & \textbf{27.06} & \textbf{29.89} & 28.58 \\
BLT & 28.80 & 28.98 & 27.92 & 26.02 & 28.68 & 27.42 \\
AU-Net & 28.64 & 28.86 & 27.54 & 26.10 & 28.42 & 27.56 \\
\midrule
Continuum Attention & \textbf{30.70} & 28.58 & \textbf{30.04} & 25.76 & 30.39 & \textbf{29.37} \\
\bottomrule
\end{tabular}
\caption{Mean HellaSwag accuracy (\%) across three seeds after deterministic byte-level prompt perturbations.}
\label{tab:large_robustness}
\end{table*}

We also evaluate the robustness in Figure \ref{fig:robustness} and Table \ref{tab:large_robustness}. In the CA evaluation, clean HellaSwag accuracy is 30.70\%; removing bytes from prompts reduces it to 28.58\%, while repeating selected bytes gives 30.04\%, antspeak 25.76\%, uppercase 30.39\%, and randomcase 29.37\%.

\subsection{CUDA Execution and Verification}
\label{sec:large_cuda}

\paragraph{Computational structure.} CA requires endpoint dot products, stable exponentials, the analytic $J_0$ and $J_1$ moments for the forward operator, and $J_2$ for analytic reverse-mode gradients. The GPU implementation evaluates causal query-prefix tiles, maintains FP32 score normalization, and uses FP16 tensor-core matrix multiplications for the principal QK and weighted-value contractions. Segment weight generation and the score vector--Jacobian product use native CUDA kernels; score scaling, mixed-precision data conversions, and gain-gradient reductions are arranged to reduce full-matrix intermediate traffic. The implementation retains $\mathcal{O}(T^2d_h)$ arithmetic and materializes intermediate attention tiles opposed to the linear-memory property of FlashAttention \cite{dao2022flashattention}.

\paragraph{Operator timings.} Under an isolated V100 test with batch $B=2$, 16 heads, $T=2{,}048$, head dimension 64, the optimized analytic implementation measures 16.60\,ms for forward and backward combined (5.01\,ms forward; 11.59\,ms backward). A separately measured conventional tiled implementation takes 23.47\,ms for the same shape; a score-and-gradient fused comparator takes 19.52\,ms in a same-run comparison. These measurements refer to the \emph{attention operator only} and cannot be equated with whole-decoder or four-GPU training throughput. Full-model FP16 inference at batch 8 is approximately $7.8\times10^4$ input bytes per second on one V100 for 2,048-byte contexts.

\paragraph{Causality and gradient checks.} We independently compare numerical output and analytical Q/K/V/gain derivatives against reference calculations for context lengths through 2,048 and test all three CUDA attention execution strategies. Future-position perturbations leave preceding attention outputs bitwise unchanged in the tested cases, while gradients from prefix outputs to future query/key/value positions are zero. Full-decoder future-token perturbations at positions around query-tile boundaries likewise yield zero observed changes in earlier logits at context length 2,048. These tests confirm causal independence across the examined configurations and context lengths.

\section{Discussion and Limitations}
\label{sec:discussion}
The experiments support the usefulness of \emph{interpolated memory retrieval} as an architectural inductive bias. Continuous integration constrains how neighboring memory positions contribute jointly, without adding a separate interpolating neural network or changing the causal decoder's projection structure. By expressing the result as nonnegative normalized weights on the original values, CA preserves conventional attention's convex aggregation semantics.

\paragraph{Prior work and novelty.} Our use of a normalized density over a continuous domain belongs to a lineage of continuous attention and integral-operator research \cite{martins2020sparsecontinuous,martins2022sparsecontinuous,vuckovic2021regularity,calvello2025continuum}. Our work makes a specific contribution by defining the trajectory directly from adjacent causal Transformer key/value vectors and deriving exact segment moment formulas and analytic gradients for autoregressive language modeling.

\paragraph{Systems tradeoffs.} Exact segment integration introduces moment evaluation and coefficient redistribution beyond a softmax normalization. Our optimized CUDA implementation enables training of a 101M-parameter CA decoder with 2,048-byte contexts on four NVIDIA V100 GPUs. The current implementation retains quadratic arithmetic and materializes intermediate attention tiles, providing opportunities for further kernel optimization.

\paragraph{Generalization.}
The results reveal domain-dependent strengths, with the largest improvements in multilingual language modeling and gains on HellaSwag and ARC-Challenge. Source-code performance highlights an opportunity to further investigate continuous memory representations across different data distributions.

\section{Conclusion}
\label{sec:conclusion}

We introduced Continuum Attention, a causal retrieval operator that exactly integrates query-dependent relevance over piecewise linear key and value trajectories. Its closed-form exponential moments enable analytical forward and backward computation while preserving normalized, nonnegative attention weights and compatibility with standard Transformer decoders.

Controlled five-seed experiments demonstrate improvements over temperature-matched softmax attention at both evaluated context lengths. Our three-seed, 101M-parameter study demonstrates a 42\% reduction in training bytes required to reach a matched training BPB threshold and superior FLORES multilingual performance, achieving a 5.8\% BPB reduction relative to the strongest baseline. The results also reveal domain-dependent performance, including weaker outcomes on source code.

These findings establish exact continuous memory integration as a practical alternative to discrete causal attention and provide a foundation for further advances in continuous retrieval mechanisms.

\paragraph{Code availability.} Metal and NVIDIA CUDA implementations, test programs, and launchers are available at \url{https://github.com/akhadangi/continuum-attention}.


\newpage
{\small
\bibliographystyle{plain}
\bibliography{references}
}

\newpage

\appendix
\renewcommand{\thefigure}{\thesection.\arabic{figure}}
\renewcommand{\thetable}{\thesection.\arabic{table}}
\renewcommand{\theequation}{\thesection.\arabic{equation}}
\setcounter{figure}{0}
\setcounter{table}{0}
\setcounter{equation}{0}

\section{Controlled Small-Model Study: Setup}
\label{app:experimental_setup}

\paragraph{Dataset and evaluation task.}
We conduct byte level language modeling on TinyStories
\cite{eldan2023tinystories}. Each text is encoded as a
sequence of UTF-8 bytes, giving a vocabulary of 256 symbols.
The model predicts the next byte using the standard
autoregressive cross entropy objective.

We construct the training corpus from the first
16 million bytes of the TinyStories training split.
Development validation uses the first 2 million bytes of
the official validation split. An additional evaluation
corpus consists of a separate 4 million byte portion of
the validation stream, disjoint from the development
validation bytes.

Our primary metric is negative log likelihood (NLL),
measured in nats per byte. We additionally report bits
per byte (BPB), computed as
\begin{equation}
    \mathrm{BPB}
    =
    \frac{\mathrm{NLL}}{\ln 2}.
    \label{eq:experiment_bpb}
\end{equation}
Lower values indicate better predictive performance.

\paragraph{Model architecture.}
All experiments use the same causal Transformer decoder
architecture \cite{vaswani2017attention}. The network
contains six Transformer blocks, model width 256, four
attention heads, and head dimension 64. Each block uses
RMS normalization \cite{zhang2019rmsnorm}, residual
connections, and a SwiGLU feedforward network with hidden
dimension 720 \cite{shazeer2020glu}.

The model employs learned absolute positional embeddings,
a vocabulary embedding of dimension 256, and an output
projection tied to the input embedding matrix. The
attention operator is the principal architectural variable
across experiments.

At context length 192, fixed temperature models contain
5,008,640 trainable parameters. At context length 512,
they contain 5,090,560 parameters. The difference arises
from the larger positional embedding table. Variants with
learned attention gains introduce 24 additional scalar
parameters, corresponding to four heads in each of six
layers.

\paragraph{Attention configurations.}
We compare eight configurations, divided into conventional
softmax attention and CA variants.

The conventional softmax controls are:
\begin{itemize}
    \item $\mathrm{B}_{0.20}$, $\mathrm{B}_{0.30}$,
    and $\mathrm{B}_{0.50}$:
    softmax attention with fixed temperatures
    $\tau\in\{0.20,0.30,0.50\}$.
    \item $\mathrm{F}$:
    softmax attention with a learned positive gain
    initialized at effective temperature $\tau=1$.
    \item $\mathrm{G}$:
    softmax attention with a learned positive gain
    initialized at effective temperature $\tau=0.30$.
\end{itemize}

The CA configurations are:
\begin{itemize}
    \item $\mathrm{S}$:
    CA with learned attention gains initialized
    at effective temperature $\tau=1$.
    \item $\mathrm{H}$:
    CA with learned attention gains initialized
    at effective temperature $\tau=0.30$.
    \item $\mathrm{P}$:
    CA with fixed attention temperature $\tau=0.30$.
\end{itemize}

For learned gain variants, every attention head has a
positive scale parameter $\gamma=\exp(\rho)$, giving
effective temperature $\tau=\exp(-\rho)$.
The fixed configurations $\mathrm{B}_{0.30}$ and
$\mathrm{P}$ therefore have identical parameter counts,
identical initial score scaling, and the same surrounding
Transformer architecture. Similarly, $\mathrm{G}$ and
$\mathrm{H}$ provide matched comparisons with learned
attention gains.

\paragraph{Optimization protocol.}
All models are trained with AdamW
\cite{loshchilov2019adamw}, using learning rate
$4\times10^{-4}$, momentum coefficients
$(\beta_1,\beta_2)=(0.9,0.95)$, weight decay $0.1$,
and gradient norm clipping at $1.0$. The learning rate
is held constant during training.

Learned attention gain parameters receive eight times
the base learning rate and no weight decay. All other
optimization settings remain identical across attention
variants.

We use five training seeds:
\[
\{1729,\ 2718,\ 3141,\ 4567,\ 6174\}.
\]

The two context configurations receive equal training
token budgets:
\begin{equation}
\begin{aligned}
    T=192:&\quad
    8\times192\times2500
    =3{,}840{,}000,\\
    T=512:&\quad
    2\times512\times3750
    =3{,}840{,}000.
\end{aligned}
\end{equation}

Here the factors correspond to batch size, sequence
length, and optimization steps. Each variant uses the
same training sample sequence for a given seed and
context length. Models are evaluated at the final
training step, with no checkpoint selection based on
validation performance.

The study contains 80 completed training runs:
eight attention configurations, five seeds, and
two context lengths.

\paragraph{Validation protocol.}
Development NLL is estimated using 128 sampled validation
batches with evaluation seed 9090, shared across all
configurations. The additional evaluation uses 256
sampled batches with evaluation seed 121212.

All comparisons use paired training seeds and identical
evaluation sampling within each context. We report
the mean NLL over five seeds and paired 95\% Student
$t$ confidence intervals for the NLL differences.
For paired differences $\delta_1,\ldots,\delta_5$,
the interval is
\begin{equation}
    \bar{\delta}
    \pm
    t_{0.975,4}
    \frac{s_{\delta}}{\sqrt{5}},
    \label{eq:experiment_ci}
\end{equation}
where $s_{\delta}$ is the sample standard deviation
of the paired differences.

These intervals characterize variation across the five
training seeds on the fixed evaluation samples. They
are reported without correction for multiple comparisons.

\subsection{Small-Model Development Validation}
\label{sec:development_results}

Table~\ref{tab:development_nll} reports mean validation
NLL for all eight attention configurations at the
two context lengths.

\begin{table}[t]
\centering
\small
\setlength{\tabcolsep}{5pt}
\begin{tabular}{l l r r}
\toprule
Model & Attention setting
& $T=192$ & $T=512$ \\
\midrule
$\mathrm{B}_{0.20}$ & Softmax, fixed $0.20$
    & 1.047540 & 1.256038 \\
$\mathrm{B}_{0.30}$ & Softmax, fixed $0.30$
    & 1.029903 & 1.174585 \\
$\mathrm{B}_{0.50}$ & Softmax, fixed $0.50$
    & 1.039986 & 1.209809 \\
$\mathrm{F}$ & Softmax, learned, init. $1$
    & 1.078360 & 1.264449 \\
$\mathrm{G}$ & Softmax, learned, init. $0.30$
    & 1.026097 & 1.200387 \\
\midrule
$\mathrm{S}$ & CA, learned, init. $1$
    & 1.038428 & 1.210424 \\
$\mathrm{H}$ & CA, learned, init. $0.30$
    & 1.013707 & 1.127689 \\
$\mathrm{P}$ & CA, fixed $0.30$
    & \textbf{1.009888} & \textbf{1.126540} \\
\bottomrule
\end{tabular}
\caption{
Mean development validation NLL in nats per byte
over five training seeds. Lower values are better.
All configurations use an equal training token budget
within each context.
}
\label{tab:development_nll}
\end{table}

The results reveal an important dependence on attention
temperature. The original CA configuration $\mathrm{S}$,
initialized at effective temperature $1$, improves upon
the correspondingly initialized softmax control
$\mathrm{F}$. However, fixed temperature softmax with
$\tau=0.30$ achieves lower NLL than $\mathrm{S}$ at both
context lengths. This motivates comparisons in which
both operators receive identical attention scaling.

Under these matched conditions, CA achieves the lowest
mean validation NLL at both context lengths.
The fixed temperature variant $\mathrm{P}$ improves
over $\mathrm{B}_{0.30}$ by
$0.020016$ nats per byte at $T=192$ and
$0.048045$ nats per byte at $T=512$.
These correspond to relative NLL reductions of
approximately $1.94\%$ and $4.09\%$, respectively.

The improvement is consistent across all five paired
seeds at both context lengths. At $T=192$, the paired
95\% confidence interval for the improvement is
$[0.013779,0.026253]$. At $T=512$, the corresponding
interval is $[0.033504,0.062587]$.

The learned gain comparison produces a similar trend.
CA variant $\mathrm{H}$ improves over softmax variant
$\mathrm{G}$ by $0.012389$ nats per byte at $T=192$
and $0.072698$ nats per byte at $T=512$.
The corresponding confidence intervals are
$[-0.004571,0.029349]$ and
$[0.030012,0.115383]$.
Although $\mathrm{H}$ wins all five paired seeds at
both lengths, the interval at $T=192$ includes zero.

\subsection{Small-Model Gain Ablation}
\label{sec:gain_ablation}

We further examine whether the attention gain must be
learned to obtain the benefits of continuous integration.

At $T=192$, fixed temperature CA $\mathrm{P}$
outperforms learned gain CA $\mathrm{H}$ by
$0.003820$ nats per byte on average. The paired
95\% confidence interval is
$[-0.003918,0.011558]$.

At $T=512$, the corresponding mean advantage
of $\mathrm{P}$ is $0.001149$ nats per byte,
with confidence interval
$[-0.014050,0.016348]$.

Neither comparison establishes a clear advantage for
learning the gain. The fixed temperature configuration
provides competitive or better mean performance in both
contexts while preserving exact parameter equality
with the fixed temperature softmax baseline.

These observations motivate the use of $\mathrm{P}$
as the primary CA configuration in the remaining
analysis. Its query adaptation arises from the
continuous relevance density and segment integration,
independently of learned temperature parameters.

\subsection{Small-Model Additional Validation}
\label{sec:holdout_results}

We evaluate the saved model checkpoints on an additional
4 million byte portion of TinyStories validation text,
disjoint from the development validation corpus.

During the original experiment, the strongest softmax
control was selected using development validation:
$\mathrm{G}$ for $T=192$ and
$\mathrm{B}_{0.30}$ for $T=512$.
These selections were frozen before the original
evaluation of $\mathrm{S}$.

Subsequently, the $\mathrm{H}$ and $\mathrm{P}$
ablations were introduced and evaluated on the same
previously examined validation portion. We therefore
treat their results as exploratory measurements.
The evaluation samples and original softmax results
are reused without retraining the controls.

Table~\ref{tab:holdout_nll} reports the mean NLL
over five seeds.

\begin{table}[t]
\centering
\small
\setlength{\tabcolsep}{5pt}
\begin{tabular}{l r r}
\toprule
Configuration
& $T=192$ & $T=512$ \\
\midrule
Selected softmax
& 1.032713 & 1.182236 \\
CA $\mathrm{S}$
& 1.046338 & 1.217520 \\
CA $\mathrm{H}$
& 1.020175 & \textbf{1.135246} \\
CA $\mathrm{P}$
& \textbf{1.016345} & 1.136495 \\
\bottomrule
\end{tabular}
\caption{
Mean exploratory NLL in nats per byte over five
training seeds on disjoint TinyStories validation
text. The selected softmax is $\mathrm{G}$ at
$T=192$ and $\mathrm{B}_{0.30}$ at $T=512$.
This evaluation portion had been examined before
the additional CA ablations were conducted.
}
\label{tab:holdout_nll}
\end{table}

Both temperature adjusted CA variants achieve lower
mean NLL than the development selected softmax
baseline at both context lengths.

For the primary fixed temperature CA variant,
the mean improvement is $0.016368$ nats per byte
at $T=192$, with paired 95\% confidence interval
$[-0.000134,0.032870]$.
All five paired seeds favor CA, although the
confidence interval includes zero.

At $T=512$, fixed temperature CA improves mean
NLL by $0.045741$ nats per byte relative to
$\mathrm{B}_{0.30}$, with confidence interval
$[0.031634,0.059848]$. The improvement occurs
in all five paired seeds.

Learned gain CA $\mathrm{H}$ similarly achieves
an improvement of $0.046990$ nats per byte at
$T=512$, with confidence interval
$[0.024375,0.069606]$ and five wins
across five seeds.

The results support a consistent performance
advantage for temperature adjusted CA at context
length 512 in this experimental setting.
At context length 192, the exploratory results
favor CA in mean NLL, with greater uncertainty
across training seeds.

Table~\ref{tab:paired_results} summarizes
the principal paired comparisons.

\begin{table*}[t]
\centering
\small
\setlength{\tabcolsep}{7pt}
\begin{tabular}{l c l r c c}
\toprule
Evaluation
& Context
& Comparison
& Mean improvement
& 95\% confidence interval
& CA wins \\
\midrule
Development
& 192 & $\mathrm{B}_{0.30}-\mathrm{P}$
& 0.020016 & $[0.013779,0.026253]$ & $5/5$ \\
Development
& 512 & $\mathrm{B}_{0.30}-\mathrm{P}$
& 0.048045 & $[0.033504,0.062587]$ & $5/5$ \\
Development
& 192 & $\mathrm{G}-\mathrm{H}$
& 0.012389 & $[-0.004571,0.029349]$ & $5/5$ \\
Development
& 512 & $\mathrm{G}-\mathrm{H}$
& 0.072698 & $[0.030012,0.115383]$ & $5/5$ \\
\midrule
Exploratory
& 192 & $\mathrm{G}-\mathrm{P}$
& 0.016368 & $[-0.000134,0.032870]$ & $5/5$ \\
Exploratory
& 192 & $\mathrm{G}-\mathrm{H}$
& 0.012538 & $[-0.004729,0.029805]$ & $4/5$ \\
Exploratory
& 512 & $\mathrm{B}_{0.30}-\mathrm{P}$
& 0.045741 & $[0.031634,0.059848]$ & $5/5$ \\
Exploratory
& 512 & $\mathrm{B}_{0.30}-\mathrm{H}$
& 0.046990 & $[0.024375,0.069606]$ & $5/5$ \\
\bottomrule
\end{tabular}
\caption{
Paired NLL differences in nats per byte.
Positive values favor CA. Confidence intervals use
Student $t$ statistics across five paired training
seeds and are unadjusted for multiple comparisons.
The exploratory comparisons reuse previously
examined validation text.
}
\label{tab:paired_results}
\end{table*}

\subsection{Small-Model Metal Performance}
\label{sec:performance_results}

We measure computational performance on Apple Silicon
using PyTorch 2.8 and the Metal Performance Shaders
backend. Conventional softmax attention uses the
PyTorch scaled dot product attention interface.
CA uses custom Metal implementations of the exact
segment integration and analytical score gradients.

At context length 192, the CA implementation uses
the \texttt{metal\_analytic} backend.
At context length 512, it uses the
\texttt{metal\_long} backend.

Training throughput includes the model forward pass,
backward pass, and optimization steps. Forward
throughput is measured separately over repeated
full sequence inference calls.

Table~\ref{tab:ca_speed} reports mean throughput
across the five training seeds.

\begin{table}[t]
\centering
\small
\setlength{\tabcolsep}{5pt}
\begin{tabular}{r l r r}
\toprule
Context & Model
& Train tok/s & Forward tok/s \\
\midrule
192 & Softmax $\mathrm{B}_{0.30}$
    & 40,547 & 153,136 \\
192 & CA $\mathrm{H}$
    & 40,010 & 129,362 \\
192 & CA $\mathrm{P}$
    & 40,715 & 131,599 \\
\midrule
512 & Softmax $\mathrm{B}_{0.30}$
    & 31,130 & 127,098 \\
512 & CA $\mathrm{H}$
    & 28,192 & 79,975 \\
512 & CA $\mathrm{P}$
    & 28,867 & 77,450 \\
\bottomrule
\end{tabular}
\caption{
Mean training and forward throughput over five
seeds on Apple Silicon. Measurements are in
tokens per second. Higher values indicate
greater throughput.
}
\label{tab:ca_speed}
\end{table}

At context length 192, fixed temperature CA achieves
training throughput comparable to conventional softmax.
At context length 512, CA $\mathrm{P}$ is approximately
$7.3\%$ slower in training throughput and $39.1\%$
slower in forward throughput than the matched
softmax control.

The larger forward performance gap reflects the
current implementation and provides an important
optimization target. These measurements characterize
the tested Metal kernels and context lengths.

\paragraph{Numerical verification.}
Before CA training, the Metal kernels are checked
against an independent CPU reference using
double precision calculations. The recorded
maximum absolute deviation for the analytical
score gradients across the experiment preflight
checks is below $6\times10^{-7}$.
All preflight checks pass at both context lengths.

\clearpage


\end{document}